\documentclass[11pt,a4paper]{article}
\usepackage[hyperref]{acl2020}
\usepackage{times}
\usepackage{latexsym}

\usepackage{microtype}
\usepackage{amsmath}
\usepackage{epsfig}
\usepackage{caption,subcaption}
\usepackage{multirow}
\usepackage{enumitem}
\usepackage{linguex}

\aclfinalcopy %
\def\aclpaperid{1512} %

\title{Recurrent Neural Network Language Models Always Learn English-Like Relative Clause Attachment}

\author{Forrest Davis \and Marten van Schijndel \\
        Department of Linguistics\\
        Cornell University\\
        \texttt{\{fd252|mv443\}@cornell.edu}}
\date{}

\begin{document}
\maketitle
\begin{abstract}

A standard approach to evaluating language models analyzes how models assign probabilities to valid versus invalid syntactic constructions (i.e.\ is a grammatical sentence more probable than an ungrammatical sentence). Our work uses ambiguous relative clause attachment to extend such evaluations to cases of multiple simultaneous valid interpretations, 
where stark grammaticality differences are absent. 
We compare model performance in English and Spanish to show
that non-linguistic biases in RNN LMs advantageously
overlap with syntactic structure in English but not Spanish. Thus, English 
models may appear to acquire human-like syntactic preferences, while models 
trained on Spanish fail to acquire comparable human-like preferences.
We conclude by relating these results to broader concerns about the relationship
between comprehension (i.e.\ typical language model use cases) and production (which generates the training data for language models), suggesting that necessary linguistic biases are not present in the
training signal at all.

\end{abstract}

\section{Introduction}

Language modeling is widely used as pretraining for many tasks involving language processing \cite{petersetal18,radfordetal18,devlinetal19}.
Since such pretraining affects so many tasks, effective evaluations to assess model quality are critical.
Researchers in the vein of the present study, typically take (pretrained) language models and ask whether those models have learned some linguistic phenomenon (e.g., subject-verb agreement). Often the task is operationalized as: do the models match some human baseline (e.g., acceptability judgments, reading times, comprehension questions) measured as humans experience this linguistic phenomenon (e.g., comparing acceptability ratings of sentences with grammatical/ungrammatical agreement). This approach tacitly assumes that the necessary linguistic biases are in the training signal and then asks whether the models learn the same abstract representations as humans given this signal. The present study casts doubt on the notion that the necessary linguistic 
biases are present in the training signal at all. 

We utilize the, now common, evaluation technique of checking whether a model assigns higher probability to grammatical sentences compared to ungrammatical sentences \cite{linzenetal16}. However, we extend 
beyond binary grammaticality. Real world applications demand that our models not only know the difference between valid and invalid sentences; they must also be able to correctly prioritize simultaneous valid interpretations \cite{lauetal17}.
In this paper, we investigate whether neural networks can in fact prioritize simultaneous interpretations in a human-like way. 
In particular, we probe the biases of neural networks for ambiguous relative clause (RC) attachments, such as the following: 

\ex. \label{ambigrel} Andrew had dinner yesterday with the \underline{nephew} of the \underline{teacher} \textit{that was divorced}.
    \citep[from][]{fernadez2003bilingual}

In \ref{ambigrel}, there are two nominals (\textit{nephew} and \textit{teacher}) that are available for modification 
by the RC (\textit{that was divorced}). We refer to attachment of the RC to the syntactically higher nominal (i.e.\ the nephew is divorced) as HIGH and attachment to the lower nominal (i.e.\ the teacher is divorced) as LOW. 

As both interpretations are equally semantically plausible when no supporting context is given, we might expect that humans choose between HIGH and LOW at chance. However, it has been widely established that English speakers tend to interpret the 
relative clause as modifying the lower nominal more often than the
higher nominal \citep[i.e.\ they have a LOW bias;\footnote{We use ``bias'' throughout this paper to refer to ``interpretation bias.'' We will return to the distinction between production bias and interpretation bias in Section \ref{discussion}.}][]{carreiras1993relative,frazier1996construal,carreiras1999esenparsing,fernadez2003bilingual}.
LOW bias is actually typologically much rarer than HIGH bias \cite{brysbaert1996modifier}. 
A proto-typical example of a language with HIGH attachment bias is Spanish
\cite[see ][]{carreiras1993relative, carreiras1999esenparsing, fernadez2003bilingual}. 

A growing 
body of literature has shown that English linguistic structures conveniently overlap
with non-linguistic biases in neural language models leading to performance advantages 
for models of English, without such models being able to learn comparable 
structures in non-English-like languages
\cite[e.g.,][]{dyer2019critical}. This, coupled with 
recent work showing that such models have a strong recency bias \cite{ravfogeletal2019}, 
suggests that one of these attachment types (LOW), will be more easily learned. Therefore, the models might appear to perform in a human-like fashion on English, while failing on the cross-linguistically more 
common attachment preference (HIGH) found in Spanish. The present study investigates 
these concerns by first establishing, via a synthetic language experiment, that recurrent neural 
network (RNN) language models (LMs) are  
capable of learning either type of attachment (Section \ref{synthetic}).
However, we then demonstrate that these models consistently exhibit
a LOW preference when trained on actual corpus data in multiple languages (English and Spanish; Sections \ref{fernandezExps}--\ref{es_replication}).

In comparing English and Spanish, we show that non-linguistic biases in RNN LMs
overlap with interpretation biases in English to appear as though the models 
have acquired English syntax, while failing to acquire minimally different
interpretation biases in Spanish. Concretely, English attachment preferences favor
the most recent nominal, which aligns with a general preference in RNN LMs for attaching to the most recent nominal. In Spanish, this general recency preference in the models remains despite a HIGH attachment 
interpretation bias in humans. 
These results raise broader questions regarding the relationship between comprehension (i.e.\ typical language model use cases) and production (which generates the training data for language models) and point to a deeper inability of RNN LMs to learn aspects of linguistic structure from raw text alone.

\section{Related Work}\label{relatedWork}

Much recent work has probed RNN LMs for their ability to represent syntactic phenomena. In particular, subject-verb agreement has been explored extensively
\cite[e.g.,][]{linzenetal16, bernardyetal2017, Enguehard17}
with results at human level performance in some cases \cite{gulordavaetal18}. However, 
additional studies have found that the models are unable 
to generalize sequential patterns to longer or shorter sequences that share the same abstract constructions \cite{trasketal2018, vanSchijndeletal2019}.
 This suggests that the learned syntactic representations are very brittle.

Despite this brittleness, RNN LMs have been claimed to exhibit human-like behavior when processing garden path constructions \cite{vanSchijndeletal2018, futrelletal2019, franketal2019}, reflexive pronouns and 
negative polarity items \cite{futrelletal2018}, and center embedding and syntactic islands \cite{wilcox2019supression, wilcox2019block}. 
There are some cases, like coordination islands, where RNN behavior is distinctly non-human \cite[see][]{wilcox2019block}, but in general this literature suggests that RNN LMs encode some type of abstract syntactic representation \cite[e.g.,][]{prasadetal2019}. Thus far though, 
the linguistic structures used to probe RNN LMs have often been those with unambiguously ungrammatical counterparts. This extends 
into the domain of semantics, where 
downstream evaluation platforms like GLUE and SuperGLUE evaluate LMs for correct vs.\ incorrect interpretations 
on tasks targeting language understanding \cite{wangetal2018, wangetal2019}.

Some recent work has relaxed this binary distinction of correct vs.\ incorrect or grammatical 
vs.\ ungrammatical. \citet{lauetal17} correlate acceptability scores generated from a LM to average human acceptability ratings, suggesting that human-like gradient syntactic knowledge can be captured by such models. \citet{futrelletal2019} also look
at gradient acceptability in both RNN LMs and humans, by focusing on alternations 
of syntactic constituency order (e.g., heavy NP shift, dative 
alternation). Their results suggest that RNN LMs acquire soft constraints on word ordering, like humans. However, the alternations 
in \citeauthor{futrelletal2019}, while varying in their degree of acceptability, maintain the same syntactic relations throughout the alternation (e.g., \textit{gave a book to Tom} and 
\textit{gave Tom a book} both preserve the fact that \textit{Tom} is the indirect object). Our work expands this line of research by probing how RNN LMs behave when multiple valid interpretations, with crucially different syntactic relations, are available within a single sentence. We find that RNN LMs do not resolve such ambiguity in a human-like way.

There are, of course, a number of other modeling approaches that exist in the current literature; the most notable of these 
being BERT \cite{devlinetal19}. These transformer models have achieved high performance on a variety of natural 
language processing tasks, however, there are a number of properties that make them less suitable to this work.
One immediate consideration 
is that of training. We are interested in the
behavior of a class of models, so we analyze the behavior of several randomly initialized models.
We do not know 
how representative BERT is of models of its same class, and training more BERT variants is immensely time consuming and environmentally
detrimental \cite{strubell2019energy}. Additionally, we are interested in probability
distributions over individual words given the preceding 
context, something that is not part of BERT's training as it takes whole sentences as input. Finally, the bidirectional 
nature of many of these models makes their representations difficult 
to compare to humans. For these reasons, we restrict our analyses to unidirectional RNN LMs. This necessarily reduces the 
generalizability of our claims. However, we still believe this work has broader implications for probing 
what aspects of linguistic representations neural networks can acquire using standard training data. 

\section{Methods}

\subsection{Experimental Stimuli} \label{stimuli}

In the present study, we compare the attachment preferences of RNN LMs to those 
established in \citet{fernadez2003bilingual}. \citeauthor{fernadez2003bilingual}
demonstrated that humans have consistent RC attachment biases using both 
self-paced reading and offline comprehension questions.
They tested both English and Spanish monolinguals 
(along with bilinguals) using parallel stimuli across the two languages,
which we adopt in the experiments in this paper.\footnote{All experimental stimuli and models used are available at \url{https://github.com/forrestdavis/AmbiAttach}}

Specifically, \citet{fernadez2003bilingual} included 24 items per language, 12 with 
a singular RC verb (\textit{was}) and 12 with a plural RC verb (\textit{were}). The 
English and Spanish stimuli are translations of each other, so they stand 
as minimal pairs for attachment preferences. Example stimuli are given below.

\ex. \label{fullstim}
    \a. Andrew had dinner yesterday with the \underline{nephew} of the teachers that was divorced. \label{fullA}
    \b. Andrew had dinner yesterday with the nephews of the \underline{teacher} that was divorced. \label{fullB}   
    \c. Andr\'e cen\'o ayer con el \underline{sobrino} de los maestros que estaba divorciado.\label{fullC}
    \d. Andr\'e cen\'o ayer con los sobrinos del \underline{maestro} que estaba divorciado. \label{fullD}
    
The underlined nominal above marks the attachment point of the relative clause (\textit{that was divorced}). 
\ref{fullA} and \ref{fullC} exhibit HIGH attachment, while \ref{fullB} and \ref{fullD} exhibit LOW attachment.
\citeauthor{fernadez2003bilingual} found that English speakers had a LOW bias, preferring \ref{fullB} over 
\ref{fullA}, while Spanish speakers had a HIGH bias, preferring  \ref{fullC} over \ref{fullD}.

We ran two experiments per language,\footnote{The vocabulary of the models was constrained to the 50K most frequent words during training. 
Out-of-vocabulary nominals in the original stimuli were replaced with semantically similar nominals. In English,
lid(s) to cover(s) and refill(s) to filler(s). In Spanish, 
sarc\'ofago(s) to ata\'ud(es), recambio(s) to sustituci\'on(es), fregadero(s) to lavabo(s), 
ba\'ul(es) to caja(s), cacerola(s) to platillo(s), and bol\'igrafo(s) to pluma(s)}
one a direct simulation of the experiment from \citet{fernadez2003bilingual}
and the other an extension (\textsc{Extended Data}), using a larger set of experimental stimuli.
The direct simulation allowed us to compare the attachment preferences for RNN LMs
to the experimental results for humans. The extension allowed us to confirm that any attachment preferences
we observed were generalizable properties of these models. 

Specifically, the \textsc{Extended Data} set of stimuli included the English and Spanish stimuli from \citet{carreiras1993relative} in addition to the stimuli from \citet{fernadez2003bilingual}, for 
a total of 40 sentences. Next, we assigned part-of-speech tags to the English and Spanish LM training data using TreeTagger
\cite{schmid1999improvements}. 
We filtered the tokens to the top 40 most frequent plural nouns, generating 
the singular forms from TreeTagger's lemmatization. We then substituted into the test sentences all combinations of distinct nouns excluding reflexives. Then we appended a relative clause with either a singular or plural verb (\textit{was/were} or \textit{estaba/estaban}).\footnote{Since the unidirectional
models are tested at the RC verb, we did not need to generate the rest of the sentence after that verb.}
Finally, each test stimulus in a pair had a LOW and HIGH attachment version for a total of 249600 sentences. An example of four sentences generated for English given 
the two nouns \textit{building} and \textit{system} is below. 

\ex. \label{en_full}
    \a. Everybody ignored the system of the buildings that was
    \b. Everybody ignored the systems of the building that was
    \c. Everybody ignored the system of the buildings that were
    \d. Everybody ignored the systems of the building that were

Not all combinations are semantically coherent; however, \citeauthor{gulordavaetal18} suggest that syntactic operations (e.g., subject-verb agreement) are still possible for RNN LMs with ``completely meaningless'' sentences
\cite[][p. 2]{gulordavaetal18}.

\subsection{RNN LM Details}\label{lm_details}

We analyzed long short-term memory networks \citep[LSTMs;][]{hochreiterschmidhuber97} throughout the present paper. 
For English, we used the English Wikipedia training data provided by \citet{gulordavaetal18}.\footnote{\url{https://github.com/facebookresearch/colorlessgreenRNNs}} 
For Spanish, we constructed a comparable training corpus
from Spanish Wikipedia following the process used by \citet{gulordavaetal18}. 
A recent dump of Spanish Wikipedia was downloaded, raw text 
was extracted using WikiExtractor,\footnote{\url{https://github.com/attardi/wikiextractor}} 
and tokenization was done using TreeTagger. A 100-million word
subset of the 
data was extracted, shuffled by sentence, and split into training (80\%) and validation (10\%) sets.%
\footnote{We also created a test partition (10\% of our data), which we did not use in this work.}
For LM training, we included the 50K most frequent words in the vocabulary, replacing the other tokens with `$\langle$UNK$\rangle$'.

\begin{table}[t]
    \centering
    \begin{tabular}{|c|c|c|c|c|c|}
        \hline
        Language & $\mu$ & $\sigma$ \\
        \hline
         Synthetic & 4.62 & 0.03 \\
         English & 51.83 & 0.96 \\ 
         Spanish & 40.80 & 0.89 \\
         \hline
    \end{tabular}
    \caption{Mean and standard deviation of LM validation perplexity for the
    synthetic models used in Section \ref{synthetic}, the
    English models used
    in Section \ref{fernandezExps}-\ref{rc}, and the Spanish
    models used in Section \ref{es_replication}}
    \label{tab:perps}
\end{table}

We used the best English model in \citet{gulordavaetal18} and trained 4 additional models with the same 
architecture\footnote{The models had 2 layers, 650 hidden/embedding units, batch size 128, dropout 0.2, and an initial learning rate of 20.} 
but different random initializations. There was no established Spanish model architecture, so we 
took the best Romance model architecture\footnote{They focused on Italian as a Romance language. The models are the same as English except the batch size is 64.} reported in \citet{gulordavaetal18} and trained 5 models. 
All models used in this work 
were trained for 40 epochs with resultant mean validation perplexities and standard deviations 
in Table \ref{tab:perps}.

\subsection{Measures}\label{measures}
We evaluated the RNN LMs using information-theoretic surprisal \cite{shannon48,hale2001probabilistic, levy2008expectation}. Surprisal is defined as the inverse log probability
assigned to each word ($w_i$) in a sentence given the preceding context. 
\[\text{surprisal}(w_i) = -\text{log}\,p(w_i|w_1 ... w_{i-1})\]

The probability is calculated by applying the softmax function to an RNN's output layer. Surprisal 
has been correlated with human processing difficulty \cite{smith2013effect,franketal15} allowing us to compare model behavior 
to human behavior. Each of the experiments done in this work looked at sentences that differed in the grammatical number of the nominals, repeated from Section \ref{stimuli} below.

\ex. \label{minimal}
    \a. Andrew had dinner yesterday with the \underline{nephew} of the teachers that was divorced. \label{minimalA}
    \b. Andrew had dinner yesterday with the nephews of the \underline{teacher} that was divorced. \label{minimalB}
    \z.
    \citep[from][]{fernadez2003bilingual}

In \ref{minimalA} the RC verb (\textit{was})
agrees with the HIGH nominal, while
in \ref{minimalB} it agrees with the LOW nominal.
As such, this
minimal pair probes the interpretation bias induced
by the relativizer (\textit{that}).

We measure the surprisal of the RC verb (\textit{was}) in both sentences of the pair. If the model has a preference for LOW attachment, then we expect that the surprisal will be smaller when the number of the final noun agrees with the number of the RC verb (e.g., surprisal \ref{minimalB} $<$ surprisal \ref{minimalA}). Concretely, for each such pair we take the difference in surprisal of the RC verb in the case of HIGH attachment \ref{minimalA} from the surprisal of the RC verb in the case of LOW attachment \ref{minimalB}. If this difference (surprisal \ref{minimalA} - surprisal \ref{minimalB}) is positive, then the LM has a LOW bias, and if 
the difference is negative, the LM has a HIGH bias.

\section{Attachment vs.\ Recency}\label{synthetic}
We begin with a proof of concept. It has been noted 
that RNN LMs have a strong recency bias \cite{ravfogeletal2019}. 
As such, it could be possible that only one type of attachment, namely LOW attachment, is learnable. To 
investigate this possibility, we followed the methodology in \citet{mccoyetal2018} and 
constructed a synthetic language to control the distribution of RC attachment in two experiments.
Our first experiment targeted the question: if all 
RC attachment is HIGH, how many RCs have to be observed in training in order for a 
HIGH bias to generalize to unseen data? Our second experiment targeted the question: what 
proportion of HIGH and LOW attachment is needed in training to learn a bias? 

Our synthetic language had RC attachment sentences and filler 
declarative sentences. The filler sentences follow the phrase
structure template given in \ref{phraseA}, while RC attachment 
sentences follow the phrase structure template given in \ref{phraseB}.

\ex. \label{phrase}
    \a.  D N (P D N) (Aux) V (D N) (P D N)  \label{phraseA}
    \b.  D N Aux V D N `of' D N `that' `was/were' V \label{phraseB}
    
Material in parentheses was optional and so was not present in all filler stimuli. That is to say, 
all filler sentences had a subject (abbreviated D N) and a verb (abbreviated V), 
with the verb being optionally transitive and 
followed by a direct object (D N).  
The subject, object, or both could be modified by a prepositional 
phrase (P D N). The subject and object could be either singular or plural, with the optional
auxiliary (Aux) agreeing in number with the subject. There were 30 nouns (N; 
60 with plural forms), 2 auxiliaries (Aux; \textit{was/were} and \textit{has/had}), 
1 determiner (D; \textit{the}), 14 verbs (V), and 4 prepositions (P). An example filler sentence is 
given in \ref{syn_filler}, and an example RC sentence is given in \ref{syn_rc}.

\ex. \label{syn_exp} 
    \a. The nephew near the children was seen by the players next to the lawyer.\label{syn_filler}
    \b. The gymnast has met the hostage of the women that was eating. \label{syn_rc}

We trained RNN LMs on our synthetic language using the same parameters as the English LMs given in 
Section \ref{lm_details}, with 120,000 unique sentences in the training corpus.
The resultant RNN LMs were 
tested on 300 sentences with ambiguous RC attachment, and we measured the surprisal 
at the RC auxiliary verb (\emph{was/were}), following the methodology given in Section \ref{measures}. 

To determine how many HIGH RCs were needed in training to learn a HIGH bias, we first constrained all the RC attachment in the 
training data to HIGH attachment. Then, we varied the proportion (in 
increments of 10 RC sentences at a time) of RC sentences to filler sentences during training.
 We trained 5 RNNs for each training configuration (i.e.\ each proportion of RCs). This experiment provided a lower bound on the 
 number of HIGH RCs needed in the training data to overcome any RNN recency bias when all RCs exhibited HIGH attachment. When as little as 0.017\% (20 sentences) of the data contained RCs with HIGH attachment, the test difference in surprisal between HIGH and LOW attachment significantly differed 
from zero ($p < 10^{-5}$, BayesFactor (BF) $> 100$),\footnote{To correct for multiple comparisons, a Bonferroni correction with $m = 6$ was used. Thus, the threshold for statistical significance was $p = 0.0083$. We also computed two-sample Bayes Factors \protect\citep[BF;][]{rouderetal09} for each statistical analysis using {\tt ttestBF} from the {\tt BayesFactor} R package \protect\cite{bayesfactor}. A Bayes Factor greater than 10 is significant evidence for the hypothesis, while 
one greater than 100 is highly significant.} with a mean difference less than zero 
($\mu = -2.24$). These results indicate that the models were able to acquire 
a HIGH bias with only 20/120000 examples of HIGH RC attachment.

In practice, we would like LMs to learn a preference even when the training data contains a mixture of HIGH and LOW attachment.
To determine the proportion of RCs that must be HIGH to learn a HIGH bias, we fixed 10\% of the training data as unambiguous 
RC attachment. Within that 10\%, we varied the proportion of HIGH and LOW attachment in 10\% 
increments (i.e.\ 0\% HIGH - 100\% LOW, 10\% HIGH - 90\% LOW, etc). Once again, 
we trained 5 models on each training configuration and tested those models on 300 test sentences, measuring the surprisal at the RC verb. 
When the training data had 50-100\% HIGH attachment,
the models preferred HIGH attachment in all the test sentences. Conversely, when the training data had 0-40\% HIGH attachment, the models preferred LOW attachment in all test sentences. 

Taken together, the results from our synthetic language experiments suggest that HIGH attachment is indeed learnable by RNN LMs. In fact, an equal proportion of HIGH and LOW attachment in the training data is all that is needed for these models to acquire a general preference for HIGH attachment (contra to the recency bias reported in the literature).

\section{English Experiments}\label{fernandezExps}

We turn now to model attachment preferences in English. We trained the models using English Wikipedia. We tested the attachment preferences of the RNN LMs using the original stimuli from \citet{fernadez2003bilingual}, 
and using a larger set of stimuli to have a better sense of model behavior on a wider range of stimuli. For 
space considerations, we only report here results of the \textsc{Extended Data} (the larger set of stimuli), but similar results hold 
for the \citet{fernadez2003bilingual} stimuli (see Supplemental Materials).

\begin{figure}[t] 
\includegraphics[width=8cm]{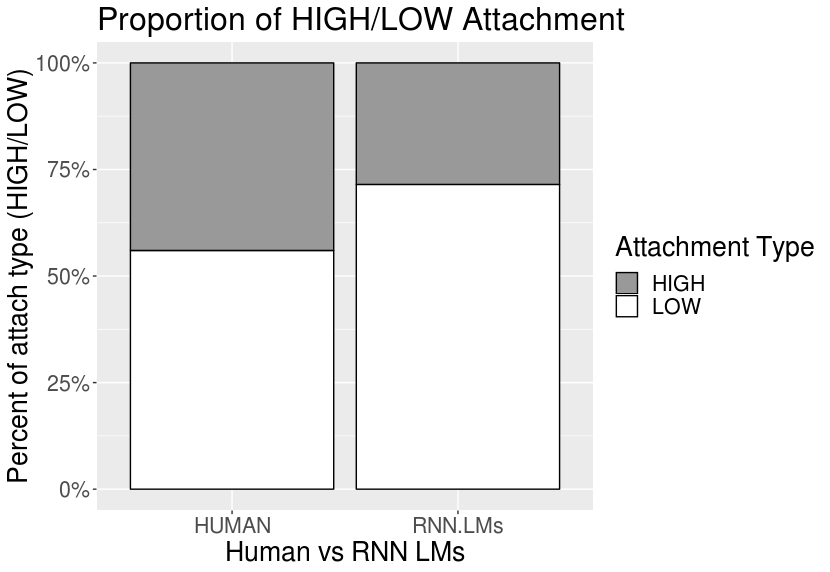}
\caption{Proportion HIGH vs LOW attachment in English. Human results from the original \citet{fernadez2003bilingual} experiment and 
RNN LM results from \textsc{Extended Data} (derived from \citet{fernadez2003bilingual} and \citet{carreiras1993relative}).}
\label{fig:en_full}
\end{figure}

In order to compare the model results with the mean human interpretation results reported by \citet{fernadez2003bilingual}, we categorically coded the model response to each item for HIGH/LOW attachment preference. 
If model surprisal for LOW attachment was less than model surprisal for HIGH attachment, the attachment was coded as LOW.
See Figure \ref{fig:en_full} for the comparison between RNNs and humans in English. 

Statistical robustness for our RNN results
was determined using the original distribution of surprisal values. Specifically, a two-tailed t-test was conducted to see if the mean difference in surprisal differed from zero (i.e.\ the model has some attachment bias). This revealed a highly significant ($p < 10^{-5}$, BF $> 100$) mean difference in surprisal of 0.77. 
This positive difference indicates that the RNN LMs have a consistent LOW bias, similar to English readers,
across models trained with differing 
random seeds.

There are two possible reasons for this patterning: (1) the models have 
learned a human-like LOW bias, or (2) the models have a recency bias that favors attachment to the lower nominal. 
These two hypotheses have overlapping predictions in English. 
The 
second hypothesis is perhaps weakened by the results of Section \ref{synthetic}, where both 
attachment types were learnable despite any recency bias. However, we know that other syntactic
attachment biases can influence RC attachment in humans \cite{scheepers2003syntactic}. It could
be that other kinds of attachment (such as prepositional phrase attachment) have varying 
proportions of attachment biases in the training data. Perhaps conflicting attachment biases across multiple constructions force the model to resort to the use of a `default' recency bias in 
cases of ambiguity. 

\section{Syntactically blocking low attachment}\label{rc}

\subsection{Stimuli}

To determine whether the behavior of the RNNs is driven by a learned attachment preference or a strong recency bias, we created
stimuli\footnote{As before, some of these stimuli are infelicitous. We do not concern ourselves with this distinction 
in the present work, given the results in \citet{gulordavaetal18}.} 
using the stimulus template described in Section \ref{stimuli} (e.g., \ref{en_full}). 
All of these stimuli 
had only the higher nominal syntactically available for attachment; the lower nominal was blocked by the addition of a relative clause:

\ex. \label{RCpair}
	\a. Everybody ignored the boy that the girls hated that was boring.\label{RCpairA}
	\b. *Everybody ignored the boys that the girl hated that was boring.\label{RCpairB}

In \ref{RCpair} only \ref{RCpairA} is grammatical. This follows because 
\textit{boy(s)} is the only nominal available for modification. In 
\ref{RCpairA}, the RC verb \textit{was} agrees in number with 
this nominal, while in \ref{RCpairB}, \textit{was} agrees in number with 
the now blocked lower nominal \textit{girl} rather than with 
\textit{boys}. For all such sentence pairs, we calculated 
the difference in surprisal between \ref{RCpairA} and \ref{RCpairB}. If their behavior is driven by a legitimate syntactic attachment preference, the models should exhibit an overwhelming HIGH bias (i.e.\ the mean difference should be less than zero). 

\begin{figure}[t] 
\includegraphics[width=8cm]{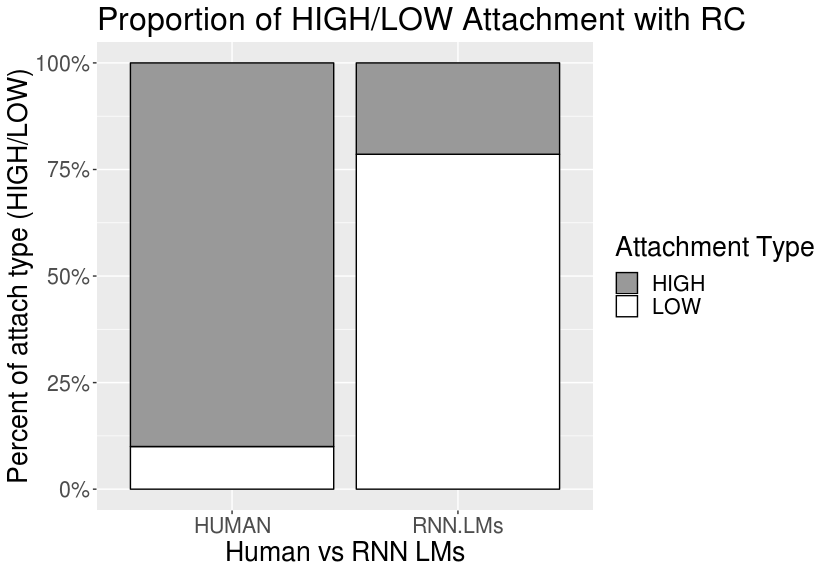}
\caption{Proportion HIGH vs LOW attachment with syntactically unavailable lower nominal. Human results estimated from \citet{linzenleonard18} and RNN LM results
from the \textsc{Extended Data} (derived from \citet{fernadez2003bilingual} and \citet{carreiras1993relative}) with the lower nominal blocked.}
\label{fig:en_RC}
\end{figure}

\subsection{Results}
As before, the differences in surprisal were calculated for each pair of experimental items. If the difference was greater than zero, the attachment was coded as LOW. The results categorically coded for HIGH/LOW attachment are given in Figure \ref{fig:en_RC}, including the results expected for humans given the pattern in \citet{linzenleonard18}.\footnote{\protect\citet{linzenleonard18} conducted experiments probing the agreement errors for subject-verb agreement with intervening RCs (and prepositional phrases). Our work is concerned with agreement between an object and its modifying RC. As such, their task serves as an approximate estimate of the errors we would expect for humans.} A two-tailed t-test was conducted to see if the mean difference in surprisal differed from zero. The results were statistically significant ($p < 10^{-5}$, BF $> 100$). The mean difference 
in surprisal was 1.15, however, suggesting that the models still had a LOW bias when the 
lower nominal was syntactically unavailable for attachment. This is in stark contrast to what one would expect if these models had 
learned the relationship between syntactic constituents and relative clause attachment. A possible alternative to the recency bias 
explanation is that RNN LMs might learn that there is a general LOW attachment bias in English and overgeneralize this pattern even in cases where one of the nominals is syntactically unavailable.

\section{The case of default HIGH bias: Spanish}\label{es_replication}

Our English analyses suggest that RNN LMs either learn a general English LOW attachment preference that they apply in all contexts, or that they have a `default' recency bias that prevents them from learning HIGH attachment preferences with more 
complex, naturalistic training data. In the case of the former, we would expect that models trained on a language whose speakers generally prefer HIGH attachment should be able to learn HIGH attachment. Spanish has a well-attested HIGH bias in humans \cite{carreiras1993relative, carreiras1999esenparsing, fernadez2003bilingual} offering a way to distinguish between 
competing recency bias and over-generalization accounts. That is, 
if the models can learn a HIGH bias when trained on Spanish data, we should be able to conclude that the general LOW bias in English is being overgeneralized by the RNNs to corner cases where HIGH bias should be preferred. 

\begin{figure}[t] 
\includegraphics[width=8cm]{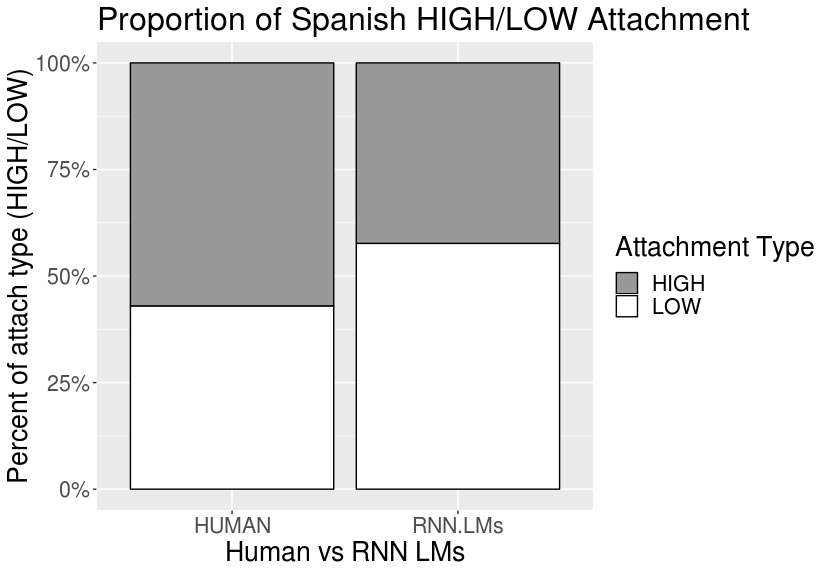}
\caption{Proportion HIGH vs LOW attachment in Spanish. Human results from the original \citet{fernadez2003bilingual} experiment and 
RNN LM results from the \textsc{Extended Data} (derived from \citet{fernadez2003bilingual} and \citet{carreiras1993relative}).}
\label{fig:es_full}
\end{figure}

\subsection{Results}
As before, the differences in surprisal were calculated for each pair of experimental items. If the difference was greater than zero, the attachment was coded as LOW. Two sample t-tests were conducted to see if the mean difference in surprisal 
differed significantly from zero for both the direct simulation of \citet{fernadez2003bilingual} and the 
\textsc{Extended Data} that included the stimuli derived from \citet{carreiras1993relative}. The results 
categorically coded for HIGH/LOW attachment for the extended stimulus set are given in Figure 
\ref{fig:es_full}, alongside the human results reported in \citet{fernadez2003bilingual}.

For the direct simulation, the mean did not differ significantly from 0 (BF $< 1/3$). 
This suggests that there is no attachment bias for the Spanish models for the stimuli from \citet{fernadez2003bilingual}, 
contrary to the human results. For the extended set of stimuli, the results were significant ($p < 10^{-5}$, BF $> 100$) with a mean difference greater than zero ($\mu = 0.211$). Thus, rather than a HIGH bias, as we would expect, the RNN LMs once again 
had a LOW bias. 

\section{Discussion}\label{discussion}

In this work, we explored the ability of RNN LMs to prioritize multiple simultaneous valid interpretations in a human-like way (as in \textit{John met the student of the teacher that was happy}).
While both LOW attachment (i.e.\ \textit{the teacher was happy}) and HIGH 
attachment (i.e.\ \textit{the student was happy}) are equally semantically plausible 
without a disambiguating context, humans have interpretation preferences 
for one attachment over the other (e.g., English speakers prefer LOW attachment and Spanish 
speakers prefer HIGH attachment).
Given the 
recent body of literature suggesting that RNN LMs have learned
abstract syntactic representations, we tested the hypothesis that
these models acquire human-like attachment preferences. 
We found that they do not.

We first used a synthetic language experiment to demonstrate that RNN LMs are capable of learning a HIGH bias when
HIGH attachment is at least as frequent as LOW attachment in the training data. 
These results suggest that any recency bias in RNN LMs is weak enough to be easily overcome by sufficient evidence of HIGH attachment.
In English, the RNNs exhibited a human-like LOW bias, but this preference persisted even in cases where LOW attachment was ungrammatical. 
To test whether the RNNs were over-learning a general LOW bias of English, 
we tested whether Spanish RNNs learned the general HIGH bias in that language. 
Once again, RNN LMs favored LOW attachment over HIGH attachment. The inability of RNN LMs to learn the Spanish HIGH attachment preference suggests that the Spanish data may not contain enough HIGH examples to learn human-like attachment preferences.

In post-hoc analyses of the Spanish Wikipedia training corpus and the AnCora Spanish newswire corpus \cite{ancora}, we find a consistent production bias towards LOW attachment among the RCs with unambiguous attachment. In Spanish Wikipedia, LOW attachment is 69\% more frequent than HIGH attachment, and in Spanish newswire data, LOW attachment is 21\% more frequent than HIGH attachment.\footnote{\url{https://github.com/UniversalDependencies/UD_Spanish-AnCora}}
This distributional bias in favor of LOW attachment does not rule out a subsequent HIGH RC bias in the models. 
It has been established in the psycholinguistic literature that attachment is learned by humans as a general abstract feature of language \cite[see][]{scheepers2003syntactic}. In other words, human syntactic representations of attachment overlap, with prepositional attachment influencing relative clause attachment, etc. 
These relationships could coalesce during training and result in an attachment preference that differs from any one structure individually. 
However, it is clear that whatever attachment biases exist in the data are insufficient for RNNs to learn a human-like attachment preference in Spanish. This provides compelling evidence that standard training data itself may systematically lack aspects of syntax relevant to performing linguistic comprehension tasks. 

We suspect that there are deep systematic issues leading to this mismatch between 
the expected distribution 
of human attachment preferences and the actual distribution of attachment in the Spanish training corpus. 
Experimental findings from psycholinguistics suggest that this issue could follow from a more general mismatch 
between language production and language comprehension. In particular, 
\citet{kehleretal2015, kehleretal2018} have provided empirical evidence that 
the production and comprehension of these structures are guided by different biases in humans. 
Production is guided by syntactic and information structural considerations 
(e.g., topic), while comprehension is influenced by those considerations plus pragmatic 
and discourse factors (e.g., coherence relations). 
As such, the biases in language production are a proper subset of those of language comprehension. As it stands 
now, RNN LMs are typically trained on production 
data (that is, the produced text in Wikipedia).\footnote{Some limited work has explored training models with human comprehension data with positive results \protect\cite{klerke-etal-2016-improving, barrett-etal-2018-sequence}.} 
Thus, they will have access to only a subset of the biases needed to learn 
human-like attachment preferences. In its strongest form, this hypothesis suggests 
that no amount of production data (i.e.\ raw text) will ever be sufficient for these models 
to generalizably pattern like humans during comprehension tasks. 

The mismatch between human interpretation biases and production biases suggested by 
this work invalidates the tacit assumption in much of the natural language processing literature that
standard, production-based training data (e.g., web text) are representative of the linguistic biases needed for 
natural language understanding and generation. 
There are phenomena, like agreement, that seem to have robust manifestations in a 
production signal, but the present work demonstrates that there are others, like 
attachment preferences, that do not. We speculate that the difference may lie in the 
inherent ambiguity in attachment, while agreement explicitly disambiguates a relation between two 
syntactic units. This discrepancy 
 is likely the reason that simply adding more data doesn't improve model 
quality \cite[e.g.,][]{vanSchijndeletal2019, bisk2020experience}. Future work needs to be done to understand more fully what biases are present in 
the data and learned
by language models.

Although our work raises questions about mismatches between human syntactic knowledge and the linguistic 
representations acquired by neural language models, it also shows that 
researchers can fruitfully use sentences with multiple interpretations 
to probe the linguistic representations acquired by those models. Before now, evaluations have focused 
on cases of unambiguous grammaticality (i.e.\ ungrammatical vs.\ grammatical). By using stimuli with multiple simultaneous valid interpretations, 
we found that evaluating models on single-interpretation sentences overestimates their ability 
to comprehend abstract syntax. 

\section*{Acknowledgments}
We would like to thank members of the NLP group and the C.Psyd lab at Cornell University, and the Altmann and Yee labs at University of Connecticut, who gave feedback on an earlier form of this work. We would also like to thank the three anonymous reviewers and 
Yonatan Belinkov.
Special thanks go to Dorit Abusch and John Whitman for 
invaluable suggestions and feedback, and Laure Thompson for comments on an earlier draft.

\bibliography{acl2020}

\begin{thebibliography}{45}
\expandafter\ifx\csname natexlab\endcsname\relax\def\natexlab#1{#1}\fi

\bibitem[{Barrett et~al.(2018)Barrett, Bingel, Hollenstein, Rei, and
  S{\o}gaard}]{barrett-etal-2018-sequence}
Maria Barrett, Joachim Bingel, Nora Hollenstein, Marek Rei, and Anders
  S{\o}gaard. 2018.
\newblock \href {https://doi.org/10.18653/v1/K18-1030} {Sequence classification
  with human attention}.
\newblock In \emph{Proceedings of the 22nd Conference on Computational Natural
  Language Learning}, pages 302--312, Brussels, Belgium. Association for
  Computational Linguistics.

\bibitem[{Bernardy and Lappin(2017)}]{bernardyetal2017}
Jean-Philippe Bernardy and Shalom Lappin. 2017.
\newblock Using deep neural networks to learn syntactic agreement.
\newblock \emph{Linguistic Issues in Language Technology (LiLT)}, 15.

\bibitem[{Bisk et~al.(2020)Bisk, Holtzman, Thomason, Andreas, Bengio, Chai,
  Lapata, Lazaridou, May, Nisnevich, Pinto, and Turian}]{bisk2020experience}
Yonatan Bisk, Ari Holtzman, Jesse Thomason, Jacob Andreas, Yoshua Bengio, Joyce
  Chai, Mirella Lapata, Angeliki Lazaridou, Jonathan May, Aleksandr Nisnevich,
  Nicolas Pinto, and Joseph Turian. 2020.
\newblock \href {http://arxiv.org/abs/2004.10151} {Experience grounds
  language}.
\newblock \emph{arXiv preprint arXiv:2004.10151}.

\bibitem[{Brysbaert and Mitchell(1996)}]{brysbaert1996modifier}
Marc Brysbaert and Don~C Mitchell. 1996.
\newblock Modifier attachment in sentence parsing: Evidence from dutch.
\newblock \emph{The Quarterly Journal of Experimental Psychology Section A},
  49(3):664--695.

\bibitem[{Carreiras and Clifton(1999)}]{carreiras1999esenparsing}
Manuel Carreiras and Charles Clifton. 1999.
\newblock Another word on parsing relative clauses: Eyetracking evidence from
  {S}panish and {E}nglish.
\newblock \emph{Memory \& Cognition}, 27(5):826--833.

\bibitem[{Carreiras and Clifton~Jr(1993)}]{carreiras1993relative}
Manuel Carreiras and Charles Clifton~Jr. 1993.
\newblock Relative clause interpretation preferences in {S}panish and
  {E}nglish.
\newblock \emph{Language and Speech}, 36(4):353--372.

\bibitem[{Devlin et~al.(2019)Devlin, Chang, Lee, and Toutanova}]{devlinetal19}
Jacob Devlin, Ming-Wei Chang, Kenton Lee, and Kristina Toutanova. 2019.
\newblock \href {https://arxiv.org/pdf/1810.04805.pdf} {{BERT}: Pre-training of
  deep bidirectional transformers for language understanding}.
\newblock In \emph{Proceedings of the 2019 Annual Conference of the North
  American Chapter of the Association for Computational Linguistics}.
  Association for Computational Linguistics.

\bibitem[{Dyer et~al.(2019)Dyer, Melis, and Blunsom}]{dyer2019critical}
Chris Dyer, G{\'a}bor Melis, and Phil Blunsom. 2019.
\newblock \href {https://arxiv.org/abs/1909.09428} {A critical analysis of
  biased parsers in unsupervised parsing}.
\newblock \emph{arXiv preprint arXiv:1909.09428}.

\bibitem[{Enguehard et~al.(2017)Enguehard, Goldberg, and Linzen}]{Enguehard17}
{\'E}mile Enguehard, Yoav Goldberg, and Tal Linzen. 2017.
\newblock \href {https://doi.org/10.18653/v1/K17-1003} {Exploring the syntactic
  abilities of {RNNs} with multi-task learning}.
\newblock In \emph{Proceedings of the 21st Conference on Computational Natural
  Language Learning (CoNLL 2017)}, pages 3--14. Association for Computational
  Linguistics.

\bibitem[{Fern\'{a}ndez(2003)}]{fernadez2003bilingual}
Eva~M. Fern\'{a}ndez. 2003.
\newblock \emph{Bilingual sentence processing: Relative clause attachment in
  {E}nglish and {S}panish}.
\newblock John Benjamins Publishing, Amsteradam.

\bibitem[{Frank and Hoeks(2019)}]{franketal2019}
Stefan~L Frank and John Hoeks. 2019.
\newblock \href {https://doi.org/10.31234/osf.io/mks5y} {The interaction
  between structure and meaning in sentence comprehension: Recurrent neural
  networks and reading times}.
\newblock \emph{PsyArXiv preprint:10.31234}.

\bibitem[{{Frank} et~al.(2015){Frank}, {Otten}, {Galli}, and
  {Vigliocco}}]{franketal15}
Stefan~L. {Frank}, Leun~J. {Otten}, Giulia {Galli}, and Gabriella {Vigliocco}.
  2015.
\newblock {The ERP response to the amount of information conveyed by words in
  sentences}.
\newblock \emph{Brain \& Language}, 140:1--11.

\bibitem[{Frazier and Clifton(1996)}]{frazier1996construal}
Lyn Frazier and Charles Clifton. 1996.
\newblock \emph{Construal}.
\newblock MIT Press, Cambridge, Mass.

\bibitem[{Futrell and Levy(2019)}]{futrelletal2019}
Richard Futrell and Roger Levy. 2019.
\newblock \href {https://www.aclweb.org/anthology/W19-0106/} {Do {RNN}s learn
  human-like abstract word order preferences?}
\newblock In \emph{Proceedings of the Society for Computation in Linguistics},
  volume~2, pages 50--59.

\bibitem[{Futrell et~al.(2018)Futrell, Wilcox, Morita, and
  Levy}]{futrelletal2018}
Richard Futrell, Ethan Wilcox, Takashi Morita, and Roger Levy. 2018.
\newblock \href {https://arxiv.org/abs/1809.01329} {{RNN}s as psycholinguistic
  subjects: {S}yntactic state and grammatical dependency}.
\newblock \emph{arXiv preprint arXiv:1809.01329}.

\bibitem[{Gulordava et~al.(2018)Gulordava, Bojanowski, Grave, Linzen, and
  Baroni}]{gulordavaetal18}
Kristina Gulordava, Piotr Bojanowski, Edouard Grave, Tal Linzen, and Marco
  Baroni. 2018.
\newblock \href {https://www.aclweb.org/anthology/N18-1108} {Colorless green
  recurrent networks dream hierarchically}.
\newblock In \emph{Proceedings of the 2018 Annual Conference of the North
  American Chapter of the Association for Computational Linguistics}.
  Association for Computational Linguistics.

\bibitem[{Hale(2001)}]{hale2001probabilistic}
John Hale. 2001.
\newblock \href {https://www.aclweb.org/anthology/N01-1021/} {A probabilistic
  earley parser as a psycholinguistic model}.
\newblock In \emph{Proceedings of the second meeting of the North American
  Chapter of the Association for Computational Linguistics on Language
  technologies}, pages 1--8. Association for Computational Linguistics.

\bibitem[{{Hochreiter} and {Schmidhuber}(1997)}]{hochreiterschmidhuber97}
Sepp {Hochreiter} and J\"{u}rgen {Schmidhuber}. 1997.
\newblock Long short-term memory.
\newblock \emph{Neural Computation}, 9(8):1735--1780.

\bibitem[{Kehler and Rohde(2015)}]{kehleretal2015}
Andrew Kehler and Hannah Rohde. 2015.
\newblock Pronominal reference and pragmatic enrichment: A bayesian account.
\newblock In \emph{CogSci}.

\bibitem[{Kehler and Rohde(2018)}]{kehleretal2018}
Andrew Kehler and Hannah Rohde. 2018.
\newblock Prominence and coherence in a bayesian theory of pronoun
  interpretation.
\newblock \emph{Journal of Pragmatics}.

\bibitem[{Klerke et~al.(2016)Klerke, Goldberg, and
  S{\o}gaard}]{klerke-etal-2016-improving}
Sigrid Klerke, Yoav Goldberg, and Anders S{\o}gaard. 2016.
\newblock \href {https://doi.org/10.18653/v1/N16-1179} {Improving sentence
  compression by learning to predict gaze}.
\newblock In \emph{Proceedings of the 2016 Conference of the North {A}merican
  Chapter of the Association for Computational Linguistics: Human Language
  Technologies}, pages 1528--1533, San Diego, California. Association for
  Computational Linguistics.

\bibitem[{Lau et~al.(2017)Lau, Clark, and Lappin}]{lauetal17}
Jey~Han Lau, Alexander Clark, and Shalom Lappin. 2017.
\newblock Grammaticality, acceptability, and probability: {A} probabilistic
  view of linguistic knowledge.
\newblock \emph{Cognitive Science}, 41:1202--1241.

\bibitem[{Levy(2008)}]{levy2008expectation}
Roger Levy. 2008.
\newblock Expectation-based syntactic comprehension.
\newblock \emph{Cognition}, 106(3):1126--1177.

\bibitem[{Linzen et~al.(2016)Linzen, Dupoux, and Goldberg}]{linzenetal16}
Tal Linzen, Emmanuel Dupoux, and Yoav Goldberg. 2016.
\newblock \href {https://www.aclweb.org/anthology/Q16-1037} {Assessing the
  ability of {LSTMs} to learn syntax-sensitive dependencies}.
\newblock \emph{Transactions of the Association for Computational Linguistics},
  4:521--535.

\bibitem[{Linzen and Leonard(2018)}]{linzenleonard18}
Tal Linzen and Brian Leonard. 2018.
\newblock \href {https://mindmodeling.org/cogsci2018/papers/0147/0147.pdf}
  {Distinct patterns of syntactic agreement errors in recurrent networks and
  humans}.
\newblock In \emph{Proceedings of the 2018 Annual Meeting of the Cognitive
  Science Society}, pages 690--695. Cognitive Science Society.

\bibitem[{McCoy et~al.(2018)McCoy, Frank, and Linzen}]{mccoyetal2018}
R~Thomas McCoy, Robert Frank, and Tal Linzen. 2018.
\newblock \href {https://arxiv.org/abs/1802.09091} {Revisiting the poverty of
  the stimulus: hierarchical generalization without a hierarchical bias in
  recurrent neural networks}.
\newblock In \emph{Proceedings of the 40th Annual Conference of the Cognitive
  Science Society}.

\bibitem[{Morey and Rouder(2018)}]{bayesfactor}
Richard~D. Morey and Jeffrey~N. Rouder. 2018.
\newblock \href {https://CRAN.R-project.org/package=BayesFactor}
  {\emph{BayesFactor: Computation of Bayes Factors for Common Designs}}.
\newblock R package version 0.9.12-4.2.

\bibitem[{Peters et~al.(2018)Peters, Neumann, Iyyer, Gardner, Clark, Lee, and
  Zettlemoyer}]{petersetal18}
Matthew~E. Peters, Mark Neumann, Mohit Iyyer, Matt Gardner, Christopher Clark,
  Kenton Lee, and Luke Zettlemoyer. 2018.
\newblock \href {https://aclweb.org/anthology/N18-1202} {Deep contextualized
  word representations}.
\newblock In \emph{Proceedings of the 2018 Annual Conference of the North
  American Chapter of the Association for Computational Linguistics}.
  Association for Computational Linguistics.

\bibitem[{Prasad et~al.(2019)Prasad, {van Schijndel}, and
  Linzen}]{prasadetal2019}
Grusha Prasad, Marten {van Schijndel}, and Tal Linzen. 2019.
\newblock \href {https://www.aclweb.org/anthology/K19-1007/} {Using priming to
  uncover the organization of syntactic representations in neural language
  models}.
\newblock In \emph{Proceedings of the 23rd Conference on Computational Natural
  Language Learning}.

\bibitem[{Radford et~al.(2018)Radford, Narasimhan, Salimans, and
  Sutskever}]{radfordetal18}
Alec Radford, Karthik Narasimhan, Tim Salimans, and Ilya Sutskever. 2018.
\newblock \href
  {https://s3-us-west-2.amazonaws.com/openai-assets/research-covers/language-unsupervised/language_understanding_paper.pdf}
  {Improving language understanding by generative pre-training}.
\newblock Technical report, OpenAI.

\bibitem[{Ravfogel et~al.(2019)Ravfogel, Goldberg, and
  Linzen}]{ravfogeletal2019}
Shauli Ravfogel, Yoav Goldberg, and Tal Linzen. 2019.
\newblock \href {https://www.aclweb.org/anthology/N19-1356/} {Studying the
  inductive biases of {RNN}s with synthetic variations of natural languages}.
\newblock In \emph{Proceedings of NAACL-HLT}.

\bibitem[{Rouder et~al.(2009)Rouder, Speckman, Sun, Morey, and
  Iverson}]{rouderetal09}
Jeffrey~N. Rouder, Paul~L. Speckman, Dongchu Sun, Richard~D. Morey, and
  Geoffrey Iverson. 2009.
\newblock Bayesian t-tests for accepting and rejecting the null hypothesis.
\newblock \emph{Psychonomic Bulletin \& Review}, 16(2):225--237.

\bibitem[{Scheepers(2003)}]{scheepers2003syntactic}
Christoph Scheepers. 2003.
\newblock Syntactic priming of relative clause attachments: Persistence of
  structural configuration in sentence production.
\newblock \emph{Cognition}, 89(3):179--205.

\bibitem[{Schmid(1999)}]{schmid1999improvements}
Helmut Schmid. 1999.
\newblock Improvements in part-of-speech tagging with an application to
  {G}erman.
\newblock In \emph{Natural language processing using very large corpora}, pages
  13--25. Springer.

\bibitem[{{Shannon}(1948)}]{shannon48}
Claude {Shannon}. 1948.
\newblock A mathematical theory of communication.
\newblock \emph{Bell System Technical Journal}, 27:379--423, 623--656.

\bibitem[{Smith and Levy(2013)}]{smith2013effect}
Nathaniel~J Smith and Roger Levy. 2013.
\newblock The effect of word predictability on reading time is logarithmic.
\newblock \emph{Cognition}, 128(3):302--319.

\bibitem[{Strubell et~al.(2019)Strubell, Ganesh, and
  McCallum}]{strubell2019energy}
Emma Strubell, Ananya Ganesh, and Andrew McCallum. 2019.
\newblock \href {https://www.aclweb.org/anthology/P19-1355/} {Energy and policy
  considerations for deep learning in {NLP}}.
\newblock In \emph{Proceedings of the 57th Annual Meeting of the Association
  for Computational Linguistics}.

\bibitem[{Taulé et~al.(2008)Taulé, Martí, and Recasens}]{ancora}
Mariona Taulé, M.~Antònia Martí, and Marta Recasens. 2008.
\newblock \href {http://www.lrec-conf.org/proceedings/lrec2008/} {{AnCora}:
  Multilevel annotated corpora for catalan and spanish}.
\newblock In \emph{Proceedings of the Sixth International Conference on
  Language Resources and Evaluation}.

\bibitem[{Trask et~al.(2018)Trask, Hill, Reed, Rae, Dyer, and
  Blunsom}]{trasketal2018}
Andrew Trask, Felix Hill, Scott~E Reed, Jack Rae, Chris Dyer, and Phil Blunsom.
  2018.
\newblock \href {https://arxiv.org/abs/1808.00508} {Neural arithmetic logic
  units}.
\newblock In \emph{Advances in Neural Information Processing Systems}, pages
  8035--8044.

\bibitem[{{van Schijndel} and Linzen(2018)}]{vanSchijndeletal2018}
Marten {van Schijndel} and Tal Linzen. 2018.
\newblock \href {https://cogsci.mindmodeling.org/2018/papers/0496/index.html}
  {Modeling garden path effects without explicit hierarchical syntax.}
\newblock In \emph{Proceedings of the 40th Annual Meeting of the Cognitive
  Science Society}.

\bibitem[{{van Schijndel} et~al.(2019){van Schijndel}, Mueller, and
  Linzen}]{vanSchijndeletal2019}
Marten {van Schijndel}, Aaron Mueller, and Tal Linzen. 2019.
\newblock \href {https://www.aclweb.org/anthology/D19-1592/} {Quantity doesn't
  buy quality syntax with neural language models}.
\newblock In \emph{Proceedings of the 2019 Conference on Empirical Methods in
  Natural Language Processing}. Association for Computational Linguistics.

\bibitem[{Wang et~al.(2019)Wang, Pruksachatkun, Nangia, Singh, Michael, Hill,
  Levy, and Bowman}]{wangetal2019}
Alex Wang, Yada Pruksachatkun, Nikita Nangia, Amanpreet Singh, Julian Michael,
  Felix Hill, Omer Levy, and Samuel Bowman. 2019.
\newblock \href {https://arxiv.org/abs/1905.00537} {{SuperGLUE}: A stickier
  benchmark for general-purpose language understanding systems}.
\newblock In \emph{Advances in Neural Information Processing Systems}, pages
  3261--3275.

\bibitem[{Wang et~al.(2018)Wang, Singh, Michael, Hill, Levy, and
  Bowman}]{wangetal2018}
Alex Wang, Amanpreet Singh, Julian Michael, Felix Hill, Omer Levy, and Samuel
  Bowman. 2018.
\newblock \href {https://www.aclweb.org/anthology/W18-5446/} {{GLUE}: A
  multi-task benchmark and analysis platform for natural language
  understanding}.
\newblock In \emph{Proceedings of the 2018 EMNLP Workshop BlackboxNLP:
  Analyzing and Interpreting Neural Networks for NLP}.

\bibitem[{Wilcox et~al.(2018)Wilcox, Levy, and Futrell}]{wilcox2019block}
Ethan Wilcox, Roger Levy, and Richard Futrell. 2018.
\newblock \href {https://arxiv.org/abs/1905.10431} {{What Syntactic Structures
  block Dependencies in RNN Language Models?}}
\newblock In \emph{Proceedings of the 41st Annual Meeting of the Cognitive
  Science Society}.

\bibitem[{Wilcox et~al.(2019)Wilcox, Levy, and Futrell}]{wilcox2019supression}
Ethan Wilcox, Roger Levy, and Richard Futrell. 2019.
\newblock \href {https://www.aclweb.org/anthology/W19-4819/} {Hierarchical
  representation in neural language models: Suppression and recovery of
  expectations}.
\newblock In \emph{Proceedings of the 2019 ACL Workshop BlackboxNLP: Analyzing
  and Interpreting Neural Networks for NLP}.

\end{thebibliography}
\bibliographystyle{acl_natbib}

\appendix

\section{\citet{fernadez2003bilingual} Replications}
\label{sec:appendix}

\subsection{English}

\begin{figure}[h!] 
\includegraphics[width=8cm]{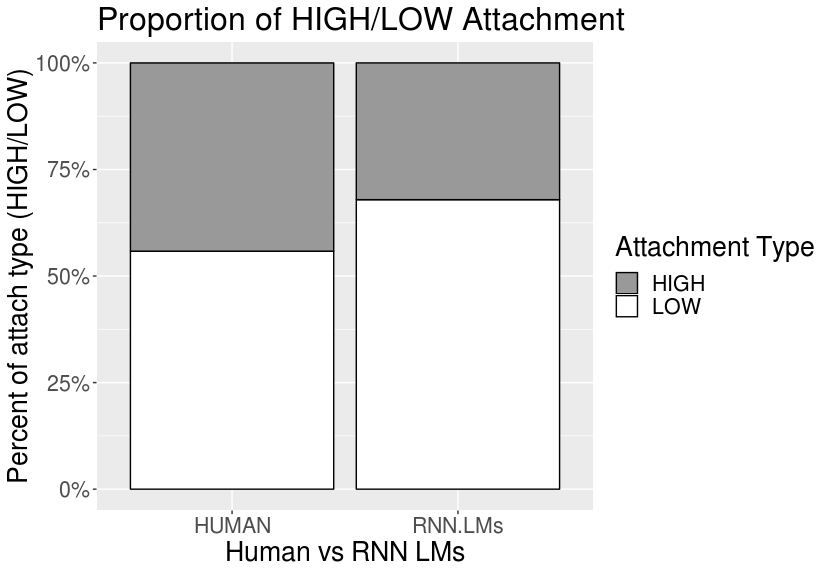}
\caption{Proportion HIGH vs LOW attachment in English. Human results from the original \citet{fernadez2003bilingual} experiment and 
RNN LM results from the stimuli 
from \citet{fernadez2003bilingual}.}
\label{fig:en_og}
\end{figure}

We compute RNN surprisal for each experimental item from \citet{fernadez2003bilingual} as detailed in Section \ref{measures} in the paper.
The results coded 
for HIGH/LOW attachment are given in Figure \ref{fig:en_og}, including the results for humans reported by \citet{fernadez2003bilingual}. While these categorical results enable easier comparison to the human results reported in the literature, statistical robustness was determined using the original distribution of surprisal values. Specifically, a two-tailed t-test was conducted to see if the mean difference in surprisal differed from zero (i.e.\ the model has some attachment bias). The result is highly significant ($p < 10^{-5}$, Bayes Factor (BF) $> 100$)
with a mean surprisal difference of $\mu = 0.66$. 
This positive difference suggests that the RNN LMs have a LOW bias, similar to English readers.

\begin{figure}[h] 
\includegraphics[width=8cm]{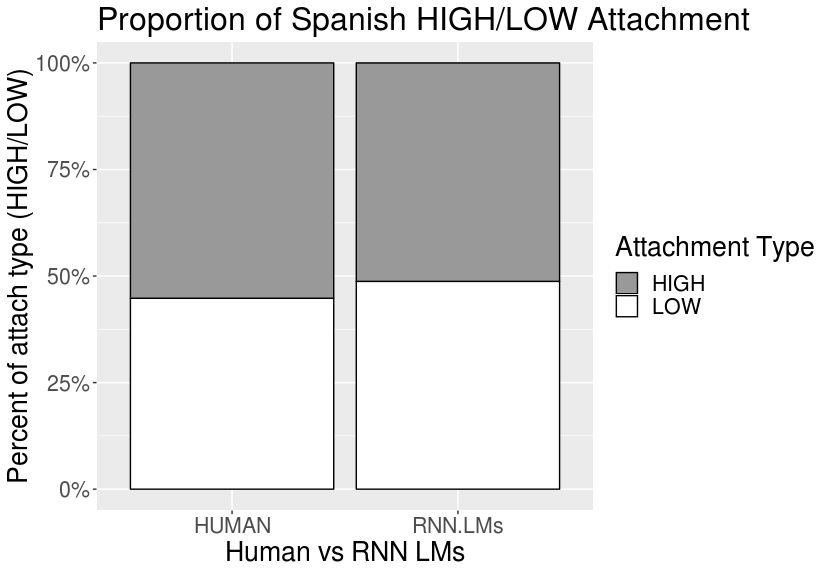}
\caption{Proportion HIGH vs LOW attachment in Spanish. Human results from the original \citet{fernadez2003bilingual} experiment and 
RNN LM results from the stimuli 
from \citet{fernadez2003bilingual}.}
\label{fig:es_og}
\end{figure}

\subsection{Spanish}

The results coded for HIGH/LOW attachment for the Spanish replication are given in Figure \ref{fig:es_og}, including the human results reported by \citet{fernadez2003bilingual}. 
The mean did not differ significantly from 0 (BF $< 1/3$). 
This suggests that there is no attachment bias for the Spanish models for the stimuli from \citet{fernadez2003bilingual}, 
contrary to the human results.

\end{document}